\pdfoutput=1
\documentclass[10pt,twocolumn,letterpaper]{article}

\usepackage[pagenumbers]{cvpr} % To force page numbers, e.g. for an arXiv version

\usepackage{graphicx}
\usepackage{amsmath}
\usepackage{amssymb}
\usepackage{booktabs}
\usepackage{multirow}
\usepackage{bbding}
\usepackage{bm}
\usepackage{makecell}
\usepackage[pagebackref,breaklinks,colorlinks]{hyperref}

\usepackage[capitalize]{cleveref}
\crefname{section}{Sec.}{Secs.}
\Crefname{section}{Section}{Sections}
\Crefname{table}{Table}{Tables}
\crefname{table}{Tab.}{Tabs.}

\def\cvprPaperID{6134} % *** Enter the CVPR Paper ID here
\def\confName{CVPR}
\def\confYear{2023}

\begin{document}

%%%%%%%%% TITLE - PLEASE UPDATE
\title{\vspace{-0.5em}Proposal-based Multiple Instance Learning for Weakly-supervised\\
Temporal Action Localization}

\author{Huan Ren\textsuperscript{1},
Wenfei Yang\textsuperscript{1},
Tianzhu Zhang\textsuperscript{1, 2, \thanks{Corresponding Author}},
Yongdong Zhang\textsuperscript{1}\\
\textsuperscript{1}University of Science and Technology of China, \textsuperscript{2}Deep Space Exploration Lab\\
{\tt\small rh\_hr\_666@mail.ustc.edu.cn, \{yangwf, tzzhang, zhyd73\}@ustc.edu.cn}
% For a paper whose authors are all at the same institution,
% omit the following lines up until the closing ``}''.
% Additional authors and addresses can be added with ``\and'',
% just like the second author.
% To save space, use either the email address or home page, not both
% \and
% Second Author\\
% Institution2\\
% First line of institution2 address\\
% {\tt\small secondauthor@i2.org}
}
\maketitle

%%%%%%%%% ABSTRACT
\begin{abstract}
Weakly-supervised temporal action localization aims to localize and recognize actions in untrimmed videos
with only video-level category labels during training.
Without instance-level annotations, most existing methods follow the Segment-based Multiple Instance Learning (S-MIL) framework,
where the predictions of segments are supervised by the labels of 
% bags. 
videos.
However, 
% the objectives about what to score are inconsistent between the training and testing stages, 
the objective for acquiring segment-level scores during training is not consistent with the target for acquiring proposal-level scores during testing, 
leading to suboptimal results.
To deal with this problem, we propose a novel Proposal-based Multiple Instance Learning (P-MIL) framework
% {\color{red} that directly classifies the proposals in the training stage}, 
that directly classifies the candidate proposals in both the training and testing stages, 
which includes three key designs:
1) a surrounding contrastive feature extraction module to suppress the discriminative short proposals by considering the surrounding contrastive information,
2) a proposal completeness evaluation module to inhibit the low-quality proposals with the guidance of the completeness pseudo labels,
and 3) an instance-level rank consistency loss to achieve robust detection by leveraging the complementarity of RGB and FLOW modalities.
Extensive experimental results on two challenging benchmarks including THUMOS14 and ActivityNet demonstrate the superior performance of our method.
Our code is available at \href{https://github.com/RenHuan1999/CVPR2023_P-MIL}{github.com/RenHuan1999/CVPR2023\_P-MIL}.

\vspace{-0.2em}

\end{abstract}

%%%%%%%%% BODY TEXT
\section{Introduction}
\label{sec:intro}
Temporal Action Localization (TAL) is one of the essential tasks in video understanding,
which aims to simultaneously discover action instances and identify their categories in untrimmed videos~\cite{cviu2017thumos,tpami2013temporal}.
TAL has recently received increasing attention from the research community due to its broad application potentials,
such as intelligent surveillance~\cite{vc2013surveillance}, video summarization~\cite{cvpr2012discovering}, highlight detection~\cite{cvpr2019highlight}, and visual question answering~\cite{emnlp2018tvqa}.
Most existing methods handle this task in a fully-supervised setting~\cite{cvpr2016msc, iccv2017ssn, cvpr2018tal-net, iccv2019bmn, cvpr2020gtad},
which requires instance-level annotations.
Despite their success, the requirements for such massive instance-level annotations limit their application in real-world scenarios.
To overcome this limitation, Weakly-supervised Temporal Action Localization (WTAL) has been widely studied
because it only requires video-level labels~\cite{cvpr2017untrimmednets, cvpr2018stpn, cvpr2019cmcs, cvpr2021ugct, cvpr2022asmloc}, which are much easier to collect.

\begin{figure}[t]
    \centering
    % Requires \usepackage{graphicx}
    \includegraphics[width=0.45\textwidth]{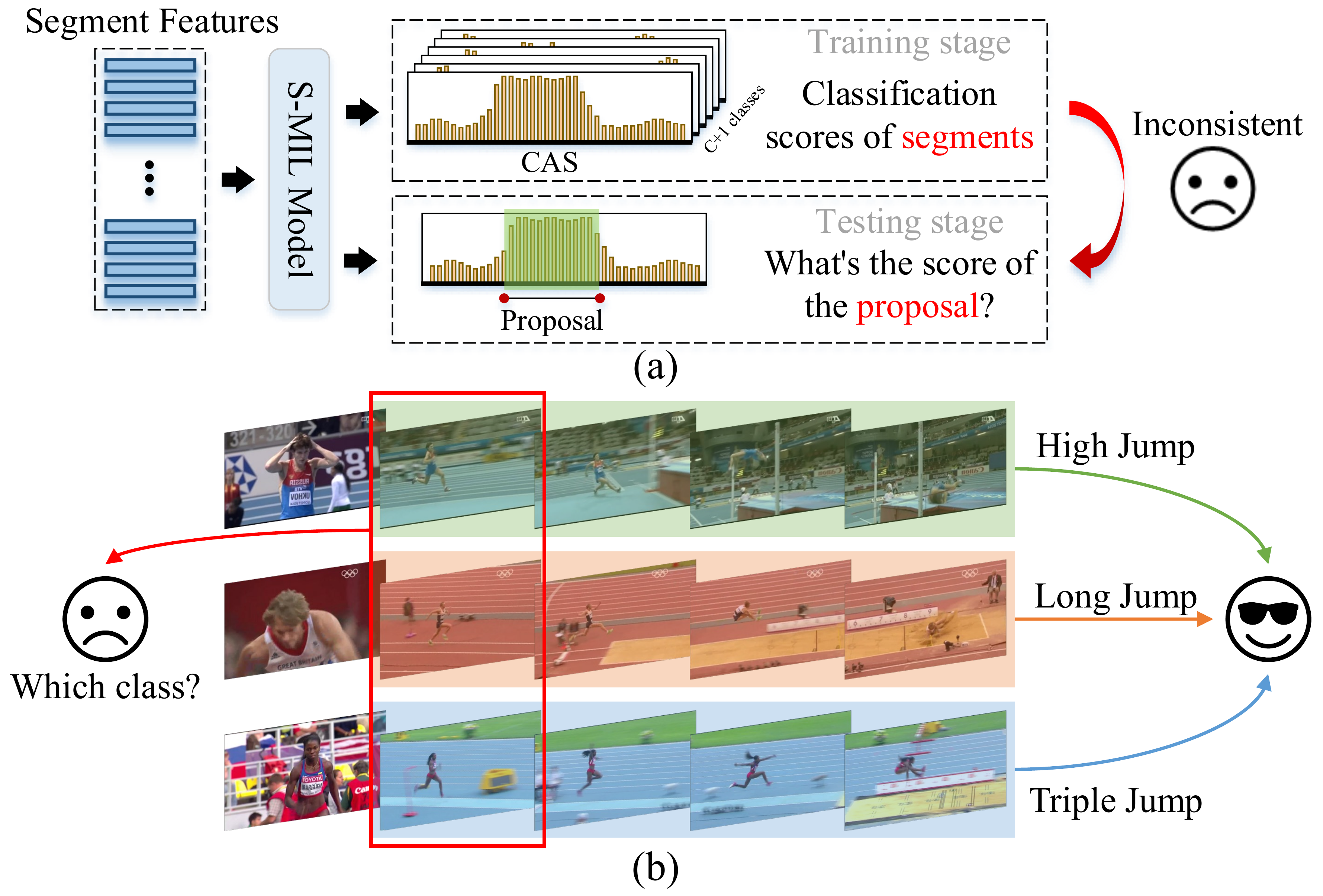}
    \vspace{-0.8em}
    \caption{
    % Motivation of the proposed method.
    Drawbacks of the Segment-based Multiple Instance Learning framework.
    (a) The objectives of the training and testing stages are inconsistent.
    (b) By watching a single running segment in the red box,
    it is difficult to tell which category it belongs to.
    }
    \label{fig:motivation}
    % \vspace{-10pt}
    % \vspace{-0.5em}
\end{figure}

Most existing WTAL methods follow the Segment-based Multiple Instance Learning (S-MIL) framework~\cite{cvpr2017untrimmednets,eccv2020EM-MIL,iccv2021facnet}, 
% to achieve temporal action localization.
where the predictions of segments are supervised by the labels of videos.
In particular, a class-agnostic attention branch is used to calculate the attention sequence,
which indicates the foreground probability of each segment.
Meanwhile, a classification branch is used to calculate the Class Activation Sequence (CAS),
which indicates the category probability of each segment.
In the training stage,
the video-level classification scores can be derived by aggregating CAS with the attention sequence,
which are then supervised by the video-level category labels.
In the testing stage, the candidate proposals are generated by thresholding the attention sequence,
and the segment-level CAS corresponding to each proposal is aggregated to score each proposal.

Despite the considerable progress achieved by these methods,
the S-MIL framework has two drawbacks.
Firstly, the objectives of the training and testing stages are inconsistent.
As shown in Figure~\ref{fig:motivation} (a),
the target is to score the action \emph{proposals} as a whole in the testing stage,
but the classifier is trained to score the \emph{segments} in the training stage.
The inconsistent scoring approach can lead to suboptimal results as shown in other weakly-supervised tasks~\cite{cvpr2016weakly, cvpr2017multiple, iccv2019towards, cvpr2019tga, tip2021local}.
Secondly, it is difficult to classify each segment alone in many cases.
As shown in Figure~\ref{fig:motivation} (b), by watching a single running segment,
it is difficult to tell whether it belongs to high jump, long jump, or triple jump.
Only by watching the entire action instance and using of the contextual information,
we can determine which category it belongs to.
%
% Secondly, the quality of localization cannot be evaluated.
% %
% Due to the discrepancy between classification and localization tasks,
% %
% it tends to focus on the most discriminative segments~\cite{MM2018step-by-step-erase, cvpr2019cmcs, cvpr2021aumn}, resulting in some incomplete localization results.
% %
% According to~\cite{iccv2019P-GCN, eccv2022react}, evaluating the localization quality of proposals can help to improve localization accuracy.
% %
% However, due to the lack of proposal-level information,
% %
% segment-based MIL methods cannot assess localization quality.
% %
% Finally, the instance-level cross-modal consistency has not been studied yet.
% %
% Previous research~\cite{eccv2020TSCN, cvpr2021ugct} indicates that cross-modal consistency between RGB and FLOW streams is essential for enhancing localization accuracy.
% %
% However, these methods only consider the “one-to-one” segment-level consistency.
% %
% In reality, each action instance corresponds to multiple segments/proposals,
% %
% and the previous approach does not take this “many-to-many” instance-level consistency into account.

Inspired by the above discussions,
we propose a novel Proposal-based Multiple Instance Learning (P-MIL) framework,
which employs a two-stage training pipeline.
In the first stage, an S-MIL model is trained and the candidate proposals are generated by
thresholding the attention sequence.
In the second stage,
the candidate proposals are classified and aggregated into video-level classification scores,
which are supervised by video-level category labels.
Since the candidate proposals are directly classified in both the training and testing stages, 
the proposed method can effectively handle the drawbacks of the S-MIL framework.
However, there are three issues that need to be considered within the P-MIL framework.
First, the model tends to focus on discriminative short proposals.
Since the training stage is mainly guided by the video-level classification,
the classifier tends to focus on the most discriminative proposals to minimize the classification loss.
To solve this problem, we propose a Surrounding Contrastive Feature Extraction module.
%
% Specifically, we extend the boundaries of the candidate proposals
% and then utilize RoIAlign~\cite{iccv2017maskrcnn} to extract the proposal features.
Specifically, we extend the boundaries of the candidate proposals
and then calculate the outer-inner contrastive features of the proposals.
By taking surrounding contrastive information into consideration,
those discriminative short proposals can be effectively suppressed.
%
% {\color{red}
% Second, the candidate proposals generated by the S-MIL approach may be incomplete.
% }
% 
% {\color{blue} 
% Second, there are many low-quality candidate proposals generated using multiple thresholds, 
% Second, the candidate proposals generated by the S-MIL approach may contain noise, 
% where a high threshold might over-segment one action instance into several incomplete proposals, 
% while a low threshold might include irrelevant background segments into the proposals.
Second, the candidate proposals generated by the S-MIL approach may be over-complete, which include irrelevant background segments. 
% }
%
In this regard, we present a Proposal Completeness Evaluation module.
%
% By treating the high-confidence proposals as \emph{pseudo instances},
% we can then acquire the completeness pseudo label of each proposal
% by computing the IoU with the pseudo instances.
Concretely speaking, we treat the high-confidence proposals as \emph{pseudo instances}, 
and then acquire the completeness pseudo label of each proposal by computing the Intersection over Union (IoU) with these pseudo instances.
Under the guidance of the completeness pseudo labels,
%
% the Proposal Completeness Evaluation module can help to inhibit the activation of low-quality proposals.
the activation of low-quality proposals can be inhibited. 
Third, due to the Non-Maximum Suppression (NMS) process in the testing stage,
the relative scores of proposals belonging to the same action instance have substantial influences on the detection results.
To learn robust relative scores,
we design an Instance-level Rank Consistency loss by
leveraging the complementarity of RGB and FLOW modalities~\cite{eccv2020TSCN, cvpr2021ugct}.
Those proposals that overlap with a given candidate proposal are considered as a cluster.
By constraining the normalized relative scores within the cluster between RGB and FLOW modalities to be consistent,
we can achieve reliable detection by discarding proposals with low relative scores in the NMS process.

To sum up, the main contributions of this paper are as follows:
(1) We propose a novel Proposal-based Multiple Instance Learning (P-MIL) framework 
% that directly classifies candidate proposals, 
% which can deal with the drawback of S-MIL.
for weakly-supervised temporal action localization, 
which can handle the drawbacks of the S-MIL framework by directly classifying the candidate proposals in both the training and testing stages.
(2) We propose three key designs (Surrounding Contrastive Feature Extraction module, Proposal Completeness Evaluation module, Instance-level Rank Consistency loss),
%{\color{red}
%which can deal with the three challenges faced by the P-MIL framework.}
%{\color{blue}
which can deal with the challenges in different stages of the P-MIL framework.
(3) Extensive experimental results on 
% two challenging benchmarks including THUMOS14 and ActivityNet
THUMOS14 and ActivityNet datasets
%
% demonstrate that the proposed framework outperforms state-of-the-art methods.
demonstrate the superior performance of the proposed framework over state-of-the-art methods.

\section{Related Work}
\label{sec:related work}
In this section, we briefly overview methods relevant to fully-supervised and weakly-supervised temporal action localization.

{\bf Fully-supervised Temporal Action Localization}. 
Temporal Action Localization (TAL) aims to simultaneously localize and identify action instances in untrimmed videos.
Similar to the development of object detection~\cite{tpami2015rcnn,nips2015faster,eccv2016ssd,eccv2020detr},
existing fully-supervised approaches can be divided into two categories:
two-stage methods~\cite{cvpr2018tal-net,cvpr2016msc,iccv2017r-c3d,iccv2017ssn,iccv2021contextloc,iccv2019P-GCN} and one-stage methods~\cite{bmvc2017end,eccv2016ssd,cvpr2019gaussian,tip2020revisiting,eccv2022tallformer,eccv2022actionformer,eccv2022efficient}.
Two-stage methods first generate the candidate proposals and then feed them into action classifiers,
by improving either the quality of proposals~\cite{cvpr2018tal-net,cvpr2016msc,iccv2019P-GCN,iccv2021contextloc} or the robustness of classifiers~\cite{iccv2017ssn,iccv2017r-c3d}.
One-stage methods can instead generate the candidate proposals and classify them simultaneously,
which have achieved remarkable performance recently by introducing Transformer architecture~\cite{eccv2022tallformer,eccv2022actionformer,eccv2022efficient}.
Despite their success, the requirements for massive and expensive instance-level annotations limit their application in real-world scenarios.

{\bf Weakly-supervised Temporal Action Localization}. 
To solve the above issue,
Weakly-supervised Temporal Action Localization (WTAL) has been widely studied~\cite{cvpr2017untrimmednets, cvpr2018stpn, cvpr2019cmcs, eccv2020TSCN, cvpr2021aumn, tpami2022ugct, cvpr2022asmloc, tip2021multi},
which requires only video-level category labels.
UntrimmedNet~\cite{cvpr2017untrimmednets} is the first to introduce a Multiple Instance Learning (MIL) framework~\cite{nips1997mil} to handle the WTAL task
by classifying segments and using a selection module to generate the action proposals.
However, due to the discrepancy between the classification and localization tasks,
the model tends to focus on the most discriminative segments.
Step-by-step~\cite{MM2018step-by-step-erase} allows the model to learn more complete localization
by gradually erasing the most discriminative segments during training.
WTALC~\cite{eccv2018wtalc} learns compact intra-class feature representations
by pulling features of the same category to be closer
while pushing those of different categories away.
By introducing a class-agnostic attention branch for foreground-background separation,
the attention-based approaches~\cite{cvpr2018stpn, cvpr2019cmcs, aaai2020background, mm2021CO2Net, cvpr2021aumn} become mainstream
due to their superior performance and the flexibility of the model architectures.
STPN~\cite{cvpr2018stpn} presents a sparsity loss on the attention sequence to capture the key foreground segments.
CMCS~\cite{cvpr2019cmcs} adopts a multi-branch architecture to discover distinctive action parts with a well-designed diversity loss.
BaS-Net~\cite{aaai2020background} and WSAL-BM~\cite{iccv2019backgroundmodel} introduce an additional background category for explicit background modeling.
CO$_2$-Net~\cite{mm2021CO2Net} designs a cross-modal attention mechanism to enhance features by filtering out task-irrelevant redundant information.
%
% AUMN~\cite{cvpr2021aumn} proposes an action unit memory network to learn action units specific classifiers.
%
More recently, there are some researches~\cite{eccv2020EM-MIL, eccv2020TSCN, cvpr2021ugct, cvpr2022rskp} that attempt to generate pseudo labels to guide the model training.
In~\cite{eccv2020EM-MIL}, the class-agnostic attention sequence and class activation sequence provide pseudo labels for each other in an expectation-maximization framework.
TSCN~\cite{eccv2020TSCN} fuses the attention sequence of RGB and FLOW modalities to generate segment-level pseudo labels,
while UGCT~\cite{cvpr2021ugct} uses the RGB and FLOW modalities to generate pseudo labels for each other by leveraging their complementarity.
Differently, ASM-Loc~\cite{cvpr2022asmloc} uses pseudo labels not only to supervise the model training
but also to enhance the segment features by leveraging the action proposals for segment-level temporal modeling.
Despite the considerable progress achieved by previous methods,
they almost all follow the \emph{Segment}-based MIL framework to achieve temporal action localization,
with inconsistent objectives between the training and testing stages.
To deal with this issue, we instead propose a novel \emph{Proposal}-based MIL framework to directly classify the candidate proposals in both the training and testing stages.

\section{Our Method}
\label{sec:method}

\begin{figure*}[ht]
    \centering
    % Requires \usepackage{graphicx}
    \includegraphics[width=1.0\linewidth]{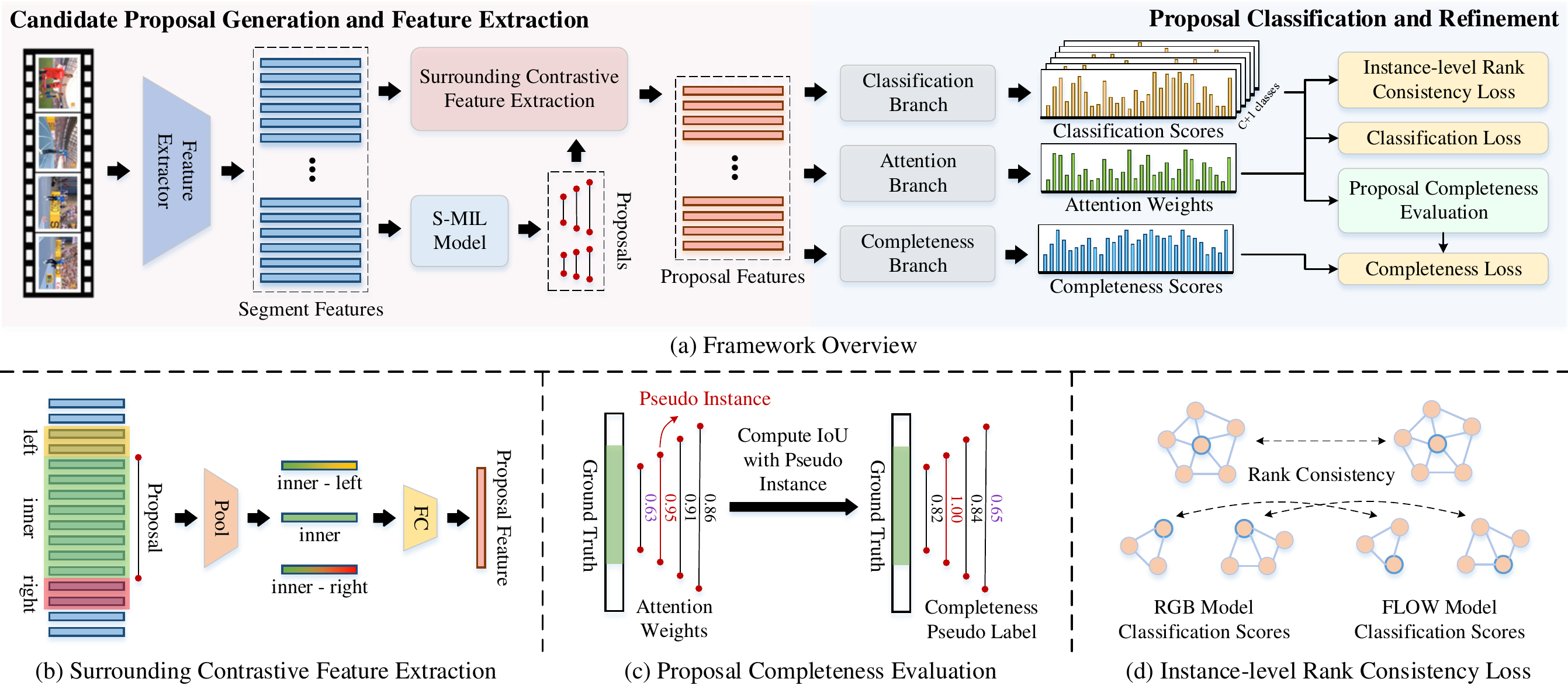}
    % \vspace{-2em}
    \caption{
    % Overall architecture of the proposed Proposal-based Multiple Instance Learning framework.
    (a) Overview of the proposed Proposal-based Multiple Instance Learning framework, which consists of candidate proposal generation, proposal feature extraction, proposal classification and refinement.
    % (b) Surrounding Contrastive Feature Extraction (SCFE) module extends the boundaries of candidate proposals and then employ RoIAlign followed by max-pooling to extract the proposal features.
    (b) The Surrounding Contrastive Feature Extraction (SCFE) module extends the boundaries of the candidate proposals and then calculates the outer-inner contrastive features of the candidate proposals.
    (c) The Proposal Completeness Evaluation (PCE) module generates the completeness pseudo label by computing the IoU with the selected pseudo instances.
    (d) The Instance-level Rank Consistency (IRC) loss constrains the normalized relative classification scores within the cluster between RGB and FLOW modalities to be consistent.
    }
    \label{fig:framework}
\end{figure*}

% %
In this section, we elaborate on the proposed Proposal-based Multiple Instance Learning framework (P-MIL) for Weakly-supervised Temporal Action Localization (WTAL),
as illustrated in Figure~\ref{fig:framework}.
%
%{\bf Task Definition}.
%
Given a video ${\bf V}$, the goal of WTAL is to predict a set of action instances $\{(c_i, s_i, e_i, q_i)\}_{i=1}^{M_p}$,
where $s_i$ and $e_i$ denote the start time and end time of the $i$-th action,
$c_i$ and $q_i$ represent the action category and the confidence score, respectively.
During training, each video ${\bf V}$ has its ground-truth video-level category label ${\bm y} \in \mathbb{R}^C$,
where $C$ represents the number of action classes.
${\bm y}(j) = 1$ if the $j$-th action presents in the video and ${\bm y}(j) = 0$ otherwise.
%
%
%{\bf Overview}.
%
The proposed P-MIL framework consists of three steps,
including candidate proposal generation (Sec.~\ref{sec:proposal-generation}),
proposal feature extraction and classification (Sec.~\ref{sec:proposal-based}),
proposal refinement (Sec.~\ref{sec:instance-level}).
The details are as follows.
%
% In the first stage, a Segment-based Multiple Instance Learning (S-MIL) model is trained to generate candidate action proposals (Sec.~\ref{sec:proposal-generation}).
% %
% In the second stage,
% there are two steps, proposal feature extraction and classification, proposal refinement.
% details are as follows.

% the candidate proposals are classified and aggregated into video-level classification scores,
% which is supervised by video-level category labels (Sec.~\ref{sec:proposal-based}).
% %
% To further mine instance-level information,
% we present a Proposal Completeness Evaluation (PCE) module to evaluate the localization completeness of proposals
% and an Instance-level Rank Consistency (IRC) loss to learn robust relative scores by leveraging the complementarity of RGB and FLOW modalities (Sec.~\ref{sec:instance-level}).

\subsection{Candidate Proposal Generation}
\label{sec:proposal-generation}
In order to generate the candidate proposals,
an S-MIL model~\cite{mm2021CO2Net} is trained.
We first divide each untrimmed video into non-overlapping 16-frame segments,
and then apply the pretrained feature extractor (\emph{e.g.} I3D~\cite{cvpr2017i3d}) to extract segment-wise features ${\bf X}_{S} \in \mathbb{R}^{T \times D}$ for both RGB and FLOW modalities,
where $T$ indicates the number of segments in the video and $D$ is the feature dimension.
%
% As the processing of RGB and FLOW modalities is parallel, without loss of generality, we omit the superscript denoting the modality in this part.
%
Following the typical two-branch architecture,
a category-agnostic attention branch is utilized to calculate the attention sequence ${\bf A} \in \mathbb{R}^{T \times 1}$
and a classification branch is used to predict the base Class Activation Sequence (CAS) ${\bf S}_{base} \in \mathbb{R}^{T \times (C+1)}$,
where the $(C+1)$-th class indicates the background~\cite{aaai2020background,iccv2019backgroundmodel}.
By multiplying ${\bf S}_{base}$ with ${\bf A}$ in the temporal dimension,
we can obtain the background suppressed CAS ${\bf S}_{supp} \in \mathbb{R}^{T \times (C+1)}$.
After that, the predicted video-level classification scores $\hat{{\bm y}}_{base}, \hat{{\bm y}}_{supp} \in \mathbb{R}^{C+1}$
are derived by applying a temporal top-$k$ aggregation strategy~\cite{aaai2020background,aaai2021HAMNet,cvpr2022asmloc} to ${\bf S}_{base}$ and ${\bf S}_{supp}$, respectively,
followed by a softmax operation.
% formalized as
% \begin{equation}\label{eq:seg_topk_orig}
% 	{\bf \hat{y}}(c) = \frac{1}{k} \mathop\mathrm{max} \limits_{\substack{{\bf p} \subset P(;, c) \\ |{\bf p}|=k}} \sum_{l=1}^{k} {\bf p}(l),
% \end{equation}
% \begin{equation}\label{eq:seg_topk_supp}
% 	{\bf \tilde{y}}(c) = \frac{1}{k} \mathop\mathrm{max} \limits_{\substack{{\bf p} \subset \bar{P}(;, c) \\ |{\bf p}|=k}} \sum_{l=1}^{k} {\bf p}(l),
% \end{equation}

Guided by the video-level category label ${\bm y}$, the classification loss is formulated as
\begin{equation}\label{eq:seg_loss_classification}
\begin{aligned}
    \mathcal{L}_{cls} = - \sum_{c=1}^{C+1} \Bigl( &{\bm y}_{base}(c) \log \hat{{\bm y}}_{base}(c) \\
    + &{\bm y}_{supp}(c) \log \hat{{\bm y}}_{supp}(c) \Bigr),
\end{aligned}
\end{equation}
where ${\bm y}_{base} = [{\bm y}, 1] \in \mathbb{R}^{C+1}$ and ${\bm y}_{supp} = [{\bm y}, 0] \in \mathbb{R}^{C+1}$, based on the assumption that background is present in all videos but filtered out by the attention sequence ${\bf A}$.
Furthermore, a sparsity loss~\cite{cvpr2018stpn} $\mathcal{L}_{norm} = \frac{1}{T} \sum_{t=1}^T |{\bf A}(t)| $
is also employed on the attention sequence ${\bf A}$ to focus on the key foreground segments.
Overall, the training objectives are as follows:
\begin{equation}\label{eq:seg_loss_total}
    \mathcal{L}_{total} = \mathcal{L}_{cls} + \lambda_{norm} \mathcal{L}_{norm},
\end{equation}
where $\lambda_{norm}$ is a balancing factor.

%{\bf Proposal Generation}.
%
With the trained S-MIL model,
we apply multiple thresholds $\theta_{act}$ on the attention sequence ${\bf A}$ to generate the candidate action proposals $P_{act} = \{ (s_i, e_i) \}_{i=1}^{M_{1}}$.
To enable our P-MIL model to learn better foreground-background separation in the training stage,
we also apply extra thresholds $\theta_{bkg}$ to generate the background proposals $P_{bkg} = \{ (s_i, e_i) \}_{i=1}^{M_{2}}$,
where the attention sequence ${\bf A}$ is below $\theta_{bkg}$.
Thus, the final candidate proposals for training are formulated as
\begin{equation}\label{eq:proposal}
    P = P_{act} + P_{bkg} = \{ (s_i, e_i) \}_{i=1}^{M},
\end{equation}
where $M = M_1 + M_2$ indicates the total number of the candidate proposals.
Note that we only use the action proposals $P_{act}$ for inference.

\subsection{Proposal Feature Extraction and Classification}
\label{sec:proposal-based}
Given the candidate proposals $P$,
previous S-MIL methods use the CAS to calculate the confidence score (\emph{e.g.} Outer-Inner Score~\cite{eccv2018autoloc}) of each proposal.
However, these indirect scoring approaches can lead to suboptimal results.
To address this problem, we propose to directly classify the candidate proposals
and aggregate them into video-level classification scores,
which are supervised by video-level category labels.

{\bf Surrounding Contrastive Feature Extraction}.
For the given candidate proposals $P$, we first extract corresponding proposal features ${\bf X}_P \in \mathbb{R}^{M \times D}$.
Since the training stage is mainly guided by the video-level classification, the classifier tends to focus on discriminative short proposals to minimize the classification loss.
To address this issue, we propose a Surrounding Contrastive Feature Extraction (SCFE) module.
Specifically, given a candidate proposal $P_i = (s_i, e_i)$, we first extend the boundary by $\alpha$ of its length on both the left and right sides, yielding three regions: \emph{left}, \emph{inner}, and \emph{right}.
For each region, we then employ RoIAlign~\cite{iccv2017maskrcnn} followed by max-pooling on the segment features ${\bf X}_S$ to extract an associated $D$-dimensional feature vector, indicated by ${\bf X}_i^{l}$, ${\bf X}_i^{n}$ and ${\bf X}_i^{r}$, respectively.
An intuitive way to obtain the proposal feature is to directly concatenate the three feature vectors and feed them into a fully connected layer.
However, inspired by AutoLoc~\cite{eccv2018autoloc}, we take a more effective approach that calculates the outer-inner contrastive features of the proposal followed by a fully connected layer, which is written as:
\begin{equation}\label{eq:contrast}
    {\bf X}_i = FC \big( \operatorname{Cat}({\bf X}_i^{n} - {\bf X}_i^{l}, {\bf X}_i^{n}, {\bf X}_i^{n} - {\bf X}_i^{r}) \big),
\end{equation}
where $\operatorname{Cat}$ denotes the concatenate operation.
By taking the surrounding contrastive information into consideration, those discriminative short proposals can be effectively suppressed.

{\bf Classification Head}.
Similar to the pipeline of the S-MIL framework, given the proposal features ${\bf X}_P$, a category-agnostic attention branch is then used to predict the attention weights ${\bf A} \in \mathbb{R}^{M \times 1}$, which indicate the foreground probability of each proposal.
Meanwhile, a classification branch is used to predict the base classification scores ${\bf S}_{base} \in \mathbb{R}^{M \times (C+1)}$ of the proposals.
By multiplying ${\bf S}_{base}$ with ${\bf A}$, we obtain the background suppressed classification scores ${\bf S}_{supp} \in \mathbb{R}^{M \times (C+1)}$.
Finally, the predicted video-level classification scores $\hat{{\bm y}}_{base}, \hat{{\bm y}}_{supp} \in \mathbb{R}^{C+1}$ are derived by applying a top-$k$ pooling followed by softmax to ${\bf S}_{base}$ and ${\bf S}_{supp}$, respectively, which are supervised by the video-level category labels.

\subsection{Proposal Refinement}
\label{sec:instance-level}
% %
% Fully-supervised TAL has instance-level labels that provide fine-grained action information,
% %
% whereas the Weakly-supervised TAL does not.
% %
% To bridge the gap between them, we mine instance-level pseudo labels for the proposed P-MIL approach.
{\bf Proposal Completeness Evaluation}.
The candidate proposals generated by the S-MIL method may be over-complete, 
which include irrelevant background segments.
In this regard, we present a Proposal Completeness Evaluation (PCE) module.
Given the candidate proposals, we use the attention weights to select the high-confidence proposals as \emph{pseudo instances},
and then obtain the completeness pseudo label of each proposal by computing the Intersection over Union (IoU) with these pseudo instances.
Formally, we first apply a threshold $\gamma \cdot max({\bf A})$ ($\gamma$ is set to $0.8$ in our case) to the attention weights ${\bf A}$ of proposals to select a set of high-confidence proposals $Q$.
Then, following the Non-Maximum Suppression (NMS) process,
we select the proposal with the highest attention weight as the pseudo instance,
remove the proposals that overlap with it from $Q$, and repeat the process until $Q$ is empty.
After that, we acquire a set of pseudo instances $G = \{ (s_i, e_i) \}_{i=1}^N$.
By computing the IoU between the candidate proposals $P$ and the pseudo instances $G$,
we can obtain an IoU matrix of $M \times N$ dimensions.
We assign the pseudo instance with the largest IoU to each proposal via taking maximum for the $N$ dimension,
and then we obtain the completeness pseudo labels ${\bm q} \in \mathbb{R}^M$ for the candidate proposals.
Under the guidance of ${\bm q}$,
a completeness branch is introduced to predict the completeness scores $\hat{{\bm q}} \in \mathbb{R}^M$ in parallel with the attention branch and the classification branch,
which can help to inhibit the activation of low-quality proposals.

{\bf Instance-level Rank Consistency}.
Due to the NMS process in the testing stage,
the relative scores of the candidate proposals belonging to the same action instance have a significant impact on the detection results.
In order to learn robust relative scores,
we design an Instance-level Rank Consistency (IRC) loss by leveraging the complementarity of RGB and FLOW modalities.
%
% In the following, we describe the IRC loss in detail.
In detail, 
we first apply a threshold $mean({\bf A})$ to the attention sequence ${\bf A}$ to eliminate the low-confidence proposals,
and the remaining proposals are denoted as $R$.
For each proposal $r$ in $R$,
those candidate proposals that overlap with it are considered as a cluster ${\mit \Omega_r}$, where $|{\mit \Omega_r}| = N_r$.
The classification scores ${\bf S}_{base}$ corresponding to this cluster are indexed from the RGB and FLOW modalities,
given as $p_{r, c}^{RGB}$ and $p_{r, c}^{FLOW}$, respectively, where $c$ indicates one of the ground truth categories.
Then the normalized relative scores within the cluster are formulated as
\begin{equation}
    D_{r, c}^{*} = \operatorname{softmax}(p_{r, c}^{*}), \forall * \in \{ RGB, FLOW \}.
\end{equation}
The Kullback-Leibler (KL) divergence is used to constrain the consistency between RGB and FLOW modalities, defined as:
\begin{equation}
\begin{aligned}
    \mathcal{L}_{IRC} = \frac{1}{|R|} \sum_{r \in R} \Big( &\operatorname{KL}(D_{r, c}^{FLOW}||D_{r, c}^{RGB}) \\
    + &\operatorname{KL}(D_{r, c}^{RGB}||D_{r, c}^{FLOW}) \Big),
\end{aligned}
\end{equation}
\begin{equation}
    \operatorname{KL}(D_{r, c}^{t}||D_{r, c}^{s}) = -\sum_{i=1}^{N_r} D_{r, c}^{t}(i) \log \frac{D_{r, c}^{s}(i)}{D_{r, c}^{t}(i)}.
\end{equation}
With the IRC loss, we can achieve reliable detection by discarding proposals with low relative scores in the NMS process.

\subsection{Network Training and Inference}
\label{sec:network}
{\bf Network Training}.
In the training stage, guided by the video-level category label ${\bm y}$, the main classification loss is formulated as
\begin{equation}\label{eq:prop_loss_classification}
\begin{aligned}
    \mathcal{L}_{cls} = - \sum_{c=1}^{C+1} \bigl( &{\bm y}_{base}(c) \log \hat{{\bm y}}_{base}(c) \\
    + &{\bm y}_{supp}(c) \log \hat{{\bm y}}_{supp}(c) \bigr),
\end{aligned}
\end{equation}
where ${\bm y}_{base} = [{\bm y}, 1] \in \mathbb{R}^{C+1}$ and ${\bm y}_{supp} = [{\bm y}, 0] \in \mathbb{R}^{C+1}$.
Moreover, with the PCE module, a completeness loss is defined as the Mean Square Error (MSE) between the completeness pseudo labels ${\bm q}$ and the predicted completeness scores $\hat{{\bm q}}$:
\begin{equation}
    \mathcal{L}_{comp} = \frac{1}{M} \sum_{i=1}^{M} \big( {\bm q}(i) - \hat{{\bm q}}(i) \big)^2.
\end{equation}
Overall, the training objective of our model is
\begin{equation}
    \label{eq:prop_loss_total}
    \mathcal{L}_{total} = \mathcal{L}_{cls} + \lambda_{comp} \mathcal{L}_{comp} + \lambda_{IRC} \mathcal{L}_{IRC},
\end{equation}
where $\lambda_{comp}$ and $\lambda_{IRC}$ are balancing hyper-parameters.

{\bf Inference}.
In the testing stage, we first apply the threshold $\theta_{cls}$ to the video-level classification scores $\hat{{\bm y}}_{supp}$ and neglect those categories below $\theta_{cls}$.
For each remaining category $c$, we score the $i$-th candidate proposal as
\begin{equation}
    {\bm s}(i) = {\bf S}_{supp}(i, c) * \hat{{\bm q}}(i).
\end{equation}
Finally, the class-wise soft-NMS~\cite{iccv2017softnms} is employed to remove the duplicate proposals.

\subsection{Discussions}
\label{sec:discussion}
In this section, we discuss the differences between the proposed method and several relevant methods, including AutoLoc~\cite{eccv2018autoloc} and CleanNet~\cite{iccv2019cleannet}.
To deal with the inconsistency between the \emph{localization} objective of the testing stage and the \emph{classification} objective of the training stage,
AutoLoc and CleanNet propose to directly predict the temporal boundaries of action instances, 
% in the training stage,
with the supervision provided by the Outer-Inner-Contrastive loss and the temporal contrast loss, respectively.
Different from these approaches,
we concentrate on another inconsistency in the S-MIL framework
about what to score between the training and testing stages.
The candidate \emph{proposals} need to be scored in the testing stage,
while the S-MIL classifier is trained to score the \emph{segments} during training.
To resolve this inconsistency,
we propose a novel Proposal-based Multiple Instance Learning framework
that directly classifies the candidate proposals in both the training and testing stages.

\begin{table*}[t]
    \vspace{-1em}
	\footnotesize
	\centering
	\caption{%
        Detection performance comparison with state-of-the-art methods on the THUMOS14 test set.
		% Note that Weakly$^+$ represents methods that utilize external weak supervisions, e.g., the numbers of action instances.
		TSN, UNT and I3D represent TSN~\cite{eccv2016tsn}, UntrimmedNet~\cite{cvpr2017untrimmednets} and I3D~\cite{cvpr2017i3d} features, respectively.
		$*$ means fusing the detection results of the S-MIL and our P-MIL model.
	}
    \vspace{-0.5em}
	\label{tab:thumos14}
	\begin{tabular}{c|c|c|ccccccc|ccc}
    \toprule
    \multirow{2.5}{*}{\bf Supervision} & \multirow{2.5}{*}{\bf Methods}   & \multirow{2.5}{*}{\bf Feature} & \multicolumn{7}{c}{\bf mAP@IoU (\%)}               & \multicolumn{3}{c}{\bf AVG mAP (\%)} \\
                                                                                               \cmidrule{4-13}
                                   &                              &                            & 0.1  & 0.2  & 0.3  & 0.4  & 0.5  & 0.6  & 0.7  & 0.1:0.5 & 0.3:0.7 & 0.1:0.7 \\
    \midrule
    \multirow{5}{*}{Fully}         
                                %    & S-CNN~\cite{cvpr2016msc}, CVPR2016                & -                        & 47.7          & 43.5          & 36.3          & 28.7          & 19.0          & -             & -             & 35.0          & -             & -             \\
                                %    & SSN~\cite{iccv2017ssn}, ICCV2017                  & -                        & 66.0          & 59.4          & 51.9          & 41.0          & 29.8          & -             & -             & 49.6          & -             & -             \\
                                   & TAL-Net~\cite{cvpr2018tal-net}, CVPR2018          & I3D                      & 59.8          & 57.1          & 53.2          & 48.5          & 42.8          & 33.8          & 20.8          & 52.3          & 39.8          & 45.1          \\
                                   & BMN~\cite{iccv2019bmn}, ICCV2019                  & TSN                      & -             & -             & 56.0          & 47.4          & 38.8          & 29.7          & 20.5          & -             & 38.5          & -             \\
                                %    & P-GCN~\cite{iccv2019P-GCN}, ICCV2019              & I3D                      & 69.5          & 67.8          & 63.6          & 57.8          & 49.1          & -             & -             & 61.6          & -             & -             \\
                                   & GTAD~\cite{cvpr2020gtad}, CVPR2020                & TSN                      & -             & -             & 54.5          & 47.6          & 40.3          & 30.8          & 23.4          & -             & 39.8          & -             \\
                                   & ContextLoc~\cite{iccv2021contextloc}, ICCV2021    & I3D                      & -             & -             & 68.3          & 63.8          & 54.3          & 41.8          & 26.2          & -             & 50.9          & -             \\
                                   & RefactorNet~\cite{cvpr2022RefactorNet}, CVPR2022  & I3D                      & -             & -             & 70.7          & 65.4          & 58.6          & 47.0          & 32.1          & -             & 54.8          & -             \\
    \midrule
    % \multirow{2}{*}{Weakly\textsuperscript{+}}
    %                                & STAR~\cite{aaai2019star}, AAAI2019                & I3D   & 68.8 & 60.0 & 48.7 & 34.7 & 23.0 & -    & -    &  47.0   &  -      &  -      \\
    %                                & 3C-Net~\cite{iccv20193cNet}, ICCV2019             & I3D   & 59.1 & 53.5 & 44.2 & 34.1 & 26.6 & -    & 8.1  &  43.5   &  -      &  -      \\
    % \midrule
    \multirow{22}{*}{Weakly}       
    % \multirow{27}{*}{Weakly}       & UntrimmedNet~\cite{cvpr2017untrimmednets}, CVPR2017        & -                        & 44.4          & 37.7          & 28.2          & 21.1          & 13.7          & -             & -             & 29.0          & -             & -             \\
								%    & Hide-and-Seek~\cite{iccv2017hide-and-seek}, ICCV2017       & -                        & 36.4          & 27.8          & 19.5          & 12.7          & 6.8           & -             & -             & 20.6          & -             & -             \\
								%    & Zhong et al.~\cite{MM2018step-by-step-erase}, MM2018       & -                        & 45.8          & 39.0          & 31.1          & 22.5          & 15.9          & -             & -             & 30.9          & -             & -             \\
								   & AutoLoc~\cite{eccv2018autoloc}, ECCV2018                   & UNT                      & -             & -             & 35.8          & 29.0          & 21.2          & 13.4          & 5.8           & -             & 21.0          & -             \\
								   & CleanNet~\cite{iccv2019cleannet}, ICCV2019                 & UNT                      & -             & -             & 37.0          & 30.9          & 23.9          & 13.9          & 7.1           & -             & 22.6          & -             \\
								   \cmidrule{2-13}
								   & STPN~\cite{cvpr2018stpn}, CVPR2018                         & I3D                      & 52.0          & 44.7          & 35.5          & 25.8          & 16.9          & 9.9           & 4.3           & 35.0          & 18.5          & 27.0          \\
								   & WTALC~\cite{eccv2018wtalc}, ECCV2018                       & I3D                      & 55.2          & 49.6          & 40.1          & 31.1          & 22.8          & -             & 7.6           & 39.8          & -             & -             \\
								   & CMCS~\cite{cvpr2019cmcs}, CVPR2019                         & I3D                      & 57.4          & 50.8          & 41.2          & 32.1          & 23.1          & 15.0          & 7.0           & 40.9          & 23.7          & 32.4          \\
								   & WSAL-BM~\cite{iccv2019backgroundmodel}, ICCV2019           & I3D                      & 60.4          & 56.0          & 46.6          & 37.5          & 26.8          & 19.6          & 9.0           & 45.5          & 27.9          & 36.6          \\
								%    & RPN~\cite{aaai2020relational}, AAAI2020                    & I3D                      & 62.3          & 57.0          & 48.2          & 37.2          & 27.9          & 16.7          & 8.1           & 46.5          & 27.6          & 36.8          \\
								%    & BaS-Net~\cite{aaai2020background}, AAAI2020                & I3D                      & 58.2          & 52.3          & 44.6          & 36.0          & 27.0          & 18.6          & 10.4          & 43.6          & 27.3          & 35.3          \\
								   & DGAM~\cite{cvpr2020DGAM}, CVPR2020                         & I3D                      & 60.0          & 54.2          & 46.8          & 38.2          & 28.8          & 19.8          & 11.4          & 45.6          & 29.0          & 37.0          \\
								   & EM-MIL~\cite{eccv2020EM-MIL}, ECCV2020                     & I3D                      & 59.1          & 52.7          & 45.5          & 36.8          & 30.5          & 22.7          & \textbf{16.4} & 44.9          & 30.4          & 37.7          \\
								%    & A2CL-PT~\cite{eccv2020A2CL-PT}, ECCV2020                   & I3D                      & 61.2          & 56.1          & 48.1          & 39.0          & 30.1          & 19.2          & 10.6          & 46.9          & 29.4          & 37.8          \\
								   & TSCN~\cite{eccv2020TSCN}, ECCV2020                         & I3D                      & 63.4          & 57.6          & 47.8          & 37.7          & 28.7          & 19.4          & 10.2          & 47.0          & 28.8          & 37.8          \\
								%    & HAM-Net~\cite{aaai2021HAMNet}, AAAI2021                    & I3D                      & 65.4          & 59.0          & 50.3          & 41.1          & 31.0          & 20.7          & 11.1          & 49.4          & 30.8          & 39.8          \\
								%    & WUM~\cite{aaai2021WSALUM}, AAAI2021                        & I3D                      & 67.5          & 61.2          & 52.3          & 43.4          & 33.7          & 22.9          & 12.1          & 51.6          & 32.9          & 41.9          \\
								%    & TS-PCA~\cite{cvpr2021tspca}, CVPR2021                      & I3D                      & 67.6          & 61.1          & 53.4          & 43.4          & 34.3          & 24.7          & 13.7          & 52.0          & 33.9          & 42.6          \\
								   & CoLA~\cite{cvpr2021cola}, CVPR2021                         & I3D                      & 66.2          & 59.5          & 51.5          & 41.9          & 32.2          & 22.0          & 13.1          & 50.3          & 32.1          & 40.9          \\
								   & AUMN~\cite{cvpr2021aumn}, CVPR2021                         & I3D                      & 66.2          & 61.9          & 54.9          & 44.4          & 33.3          & 20.5          & 9.0           & 52.1          & 32.4          & 41.5          \\
								   & UGCT~\cite{cvpr2021ugct}, CVPR2021                         & I3D                      & 69.2          & 62.9          & 55.5          & 46.5          & 35.9          & 23.8          & 11.4          & 54.0          & 34.6          & 43.6          \\
								   & CO$_2$-Net~\cite{mm2021CO2Net}, MM2021                     & I3D                      & 70.1          & 63.6          & 54.5          & 45.7          & 38.3          & 26.4          & 13.4          & 54.4          & 35.7          & 44.6          \\
								   & D2-Net~\cite{iccv2021d2net}, ICCV2021                      & I3D                      & 65.7          & 60.2          & 52.3          & 43.4          & 36.0          & -             & -             & 51.5          & -             & -             \\
								   & FAC-Net~\cite{iccv2021facnet}, ICCV2021                    & I3D                      & 67.6          & 62.1          & 52.6          & 44.3          & 33.4          & 22.5          & 12.7          & 52.0          & 33.1          & 42.2          \\
                                   & FTCL~\cite{cvpr2022ftcl}, CVPR2022                         & I3D                      & 69.6          & 63.4          & 55.2          & 45.2          & 35.6          & 23.7          & 12.2          & 53.8          & 34.4          & 43.6          \\
                                   & RSKP~\cite{cvpr2022rskp}, CVPR2022                         & I3D                      & 71.3          & 65.3          & 55.8          & 47.5          & 38.2          & 25.4          & 12.5          & 55.6          & 35.9          & 45.1          \\
                                   & ASM-Loc~\cite{cvpr2022asmloc}, CVPR2022                    & I3D                      & 71.2          & 65.5          & 57.1          & 46.8          & 36.6          & 25.2          & 13.4          & 55.4          & 35.8          & 45.1          \\
                                   & DCC~\cite{cvpr2022dcc}, CVPR2022                           & I3D                      & 69.0          & 63.8          & 55.9          & 45.9          & 35.7          & 24.3          & 13.7          & 54.1          & 35.1          & 44.0          \\
                                   \cmidrule{2-13}
                                %    & \textbf{baseline}                                          & I3D                      & 68.1          & 63.1          & 54.1          & 45.6          & 37.2          & 25.0          & 12.4          & 53.6          & 34.9          & 43.6          \\
                                   & \textbf{ours}                                   			& I3D                      & 70.9          & 66.6          & 57.8          & 48.6          & 39.8          & 27.1          & 14.4          & 56.8          & 37.5          & 46.5          \\
                                   & \textbf{ours$^*$}                            				& I3D                      & \textbf{71.8} & \textbf{67.5} & \textbf{58.9} & \textbf{49.0} & \textbf{40.0} & \textbf{27.1} & 15.1          & \textbf{57.4} & \textbf{38.0} & \textbf{47.0} \\
    \bottomrule
	\end{tabular}
    \vspace{-0.5em}
\end{table*}

\section{Experiment}
\label{sec:experiment}

\subsection{Datasets and Evaluation Metrics}
\label{sec:dataset and metric}
{\bf Datasets}.
%
% We evaluate our method on two benchmark datasets including THUMOS14~\cite{cviu2017thumos} and ActivityNet~\cite{cvpr2015activitynet}.
% %
% In {\bf THUMOS14} dataset, there are 200 validation videos and 213 test videos belonging to 20 action categories.
% %
% Follow previous works~\cite{cvpr2018stpn,eccv2018wtalc,aaai2020background}, we use the validation set for training and the test set for evaluation.
% %
% ActivityNet dataset includes two versions, ActivityNet1.2 and ActivityNet1.3.
% %
% {\bf ActivityNet1.2} dataset consists of 4819 training videos, 2383 validation videos and 2480 testing videos belonging to 100 action categories.
% %
% And {\bf ActivityNet1.3} contains 10024 training videos, 4926 validation videos and 5044 testing videos belonging to 200 action categories.
% %
% Follow previous works~\cite{iccv2019cleannet,cvpr2021aumn,cvpr2021ugct}, we use the training set for training and the validation set for evaluation.
%
We evaluate our method on two benchmark datasets including THUMOS14~\cite{cviu2017thumos} and ActivityNet~\cite{cvpr2015activitynet}.
{\bf THUMOS14} dataset contains 200 validation videos and 213 testing videos from 20 categories. 
Following previous works~\cite{cvpr2018stpn,eccv2018wtalc,aaai2020background}, we use the validation set for training and the testing set for evaluation.
{\bf ActivityNet} dataset includes two versions, ActivityNet1.2 and ActivityNet1.3, 
with 9,682 videos from 100 categories and 19,994 videos from 200 categories, respectively. 
The training, validation and testing sets are divided from ActivityNet dataset by the ratio of 2:1:1. 
%
% Since the annotations for the testing set are not released, 
Following previous works~\cite{iccv2019cleannet,cvpr2021aumn,cvpr2021ugct}, we use the training set for training and the validation set for evaluation.

{\bf Evaluation Metrics}.
In this work, we evaluate the localization performance with the mean Average Precision (mAP) values at different Intersection over Union (IoU) thresholds,
following the standard evaluation protocol~\footnote{\href{http://github.com/activitynet/ActivityNet}{http://github.com/activitynet/ActivityNet}}.

\subsection{Implementation Details}
\label{sec:implementation details}
{\bf Network Architecture}.
We employ the I3D~\cite{cvpr2017i3d} networks pretrained on Kinetics-400~\cite{cvpr2017i3d} for feature extraction.
The dimension $D$ of the extracted segment-wise features is 1024.
%
% For a fair comparison, we fix the pre-trained parameters without fine-tuning.
%
Optical flow frames are extracted from RGB frames using the TV-L1~\cite{jprs2007tv-l1} algorithm.
The category-agnostic attention branch is implemented by two fully-connected layers followed by a sigmoid activation function,
which is the same as the completeness branch.
The classification branch consists of two fully-connected layers.

{\bf Hyper-parameters Setting}.
%
% Our method is trained using the Adam~\cite{iclr2014adam} optimizer with a mini-batch size of 10.
%
% The learning rate is $5 \times 10^{-5}$ and $3 \times 10^{-5}$ for THUMOS14 and ActivityNet, respectively.
Our method is trained using the Adam~\cite{iclr2014adam} optimizer with the learning rate of $5 \times 10^{-5}$ and the mini-batch size of 10.
The extended ratio $\alpha$ is set to 0.25. 
% as recommended by AutoLoc~\cite{eccv2018autoloc}.
%
The RoI size of RoIAlign~\cite{iccv2017maskrcnn} is 2, 8, 2 for the left, inner and right region, respectively.
The loss function weights 
% $\lambda_{norm}$ = 0.5 in Equation~\eqref{eq:seg_loss_total} and 
$\lambda_{comp}$ = 20, $\lambda_{IRC}$ = 2 in Equation~\eqref{eq:prop_loss_total}.
Since the attention weights are less reliable in the early training stage,
we multiply the loss function $\mathcal{L}_{comp}$ and $\mathcal{L}_{IRC}$ with a time-varying parameter as employed in UGCT~\cite{cvpr2021ugct},
which is gradually increased to 1. 
% during the training.
%
For the candidate proposal generation, the thresholds $\theta_{act}$ = [0.1:0.1:0.9] and $\theta_{bkg}$ = [0.3:0.2:0.7].
During inference, the video-level classification threshold $\theta_{cls}$ is set to 0.2.
% During inference, the threshold $\theta_{cls}$ is 0.2.
%
% And we use the soft-NMS~\cite{iccv2017softnms} with a threshold 0.3 to remove highly overlapping proposals.

\begin{figure*}[t]
	% \vspace{-1em}
	\begin{minipage}[!t]{\linewidth}
	% \vspace{-0.8em}
	\centering
	\footnotesize
		\begin{minipage}[t]{0.48\linewidth}
			\makeatletter\def\@captype{table}
				\centering
				% \footnotesize
				\caption{%
				Detection performance comparison with state-of-the-art methods on the ActivityNet1.2 validation set.
				AVG represents the average mAP at IoU thresholds 0.5:0.05:0.95.
				$*$ means fusing the detection results of the S-MIL model and our  P-MIL model.
				}
				\label{tab:activitynet1.2}
				\vspace{-0.5em}
				\begin{tabular}{c|cccc}
				\toprule
				\multirow{2.5}{*}{\bf Methods}                          & \multicolumn{4}{c}{\bf mAP@IoU (\%)}  \\
																	\cmidrule{2-5}
																	& 0.5           & 0.75          & 0.95         & AVG           \\
				\midrule
				% UntrimmedNet~\cite{cvpr2017untrimmednets}, CVPR2017 & 7.4           & 3.9           & 1.2          & 3.6           \\
				% Zhong et al.~\cite{MM2018step-by-step-erase}, MM2018& 27.5          & 14.7          & 2.9          & 15.6          \\
				% AutoLoc~\cite{eccv2018autoloc}, ECCV2018            & 27.3          & 15.1          & 3.3          & 16.0          \\
				WTALC~\cite{eccv2018wtalc}, ECCV2018                & 37.0          & 14.6          & -            & 18.0          \\
				CMCS~\cite{cvpr2019cmcs}, CVPR2019                  & 36.8          & 22.0          & 5.6          & 22.4          \\
				% TSM~\cite{iccv2019tsm}, ICCV2019                    & 28.3          & 17.0          & 3.5          & 17.1          \\
				% CleanNet~\cite{iccv2019cleannet}, ICCV2019          & 37.1          & 20.3          & 5.0          & 21.6          \\
				% 3C-Net~\cite{iccv20193cNet}, ICCV2019               & 37.2          & 23.7          & -            & 21.7          \\
				% RPN~\cite{aaai2020relational}, AAAI2020             & 37.6          & 23.9          & 5.4          & 23.3          \\
				BaS-Net~\cite{aaai2020background}, AAAI2020         & 38.5          & 24.2          & 5.6          & 24.3          \\
				TCAM~\cite{cvpr2020TCAM} CVPR2020                   & 40.0          & 25.0          & 4.6          & 24.6          \\
				DGAM~\cite{cvpr2020DGAM}, CVPR2020                  & 41.0          & 23.5          & 5.3          & 24.4          \\
				EM-MIL~\cite{eccv2020EM-MIL}, ECCV2020              & 37.4          & 23.1          & 2.0          & 20.3          \\
				TSCN~\cite{eccv2020TSCN}, ECCV2020                  & 37.6          & 23.7          & 5.7          & 23.6          \\
				% HAM-Net~\cite{aaai2021HAMNet}, AAAI2021             & 41.0          & 24.8          & 5.3          & 25.1          \\
				% ACSNet~\cite{aaai2021acsnet}, AAAI2021              & 40.1          & 26.1          & \textbf{6.8} & 26.0          \\
				% WUM~\cite{aaai2021WSALUM}, AAAI2021                 & 41.2          & 25.6          & \textbf{6.0} & 25.9          \\
				% ASL~\cite{cvpr2021ASL}, CVPR2021                    & 40.2          & -             & -            & 25.8          \\
				CoLA~\cite{cvpr2021cola}, CVPR2021                  & 42.7          & 25.7          & 5.8          & 26.1          \\
				AUMN~\cite{cvpr2021aumn}, CVPR2021                  & 42.0          & 25.0          & 5.6          & 25.5          \\
				UGCT~\cite{cvpr2021ugct}, CVPR2021                  & 41.8          & 25.3          & \textbf{5.9} & 25.8          \\
				CO$_2$-Net~\cite{mm2021CO2Net}, MM2021              & 43.3          & \textbf{26.3} & 5.2          & 26.4          \\
				D2-Net~\cite{iccv2021d2net}, ICCV2021               & 42.3          & 25.5          & 5.8          & 26.0          \\
				\midrule
				% baseline                                            & 41.6          & 24.7          & 5.4          & 25.4          \\
				% \textbf{ours}                                     	& 42.4          & 24.8          & 4.9          & 25.2          \\
				% \textbf{ours$^*$}                              		& \textbf{44.1} & 26.1          & 5.4          & \textbf{26.4} \\
				\textbf{ours}                                     	& 42.2          & 25.0          & 4.9          & 25.5          \\
				\textbf{ours$^*$}                              		& \textbf{44.2} & 26.1          & 5.3          & \textbf{26.5} \\
				\bottomrule
				\end{tabular}
				\vspace{-1em}
		\end{minipage}
		\hspace{3mm}
		\begin{minipage}[t]{0.48\linewidth}
			\makeatletter\def\@captype{table}
				\centering
				% \footnotesize
				\caption{%
				Detection performance comparison with state-of-the-art methods on the ActivityNet1.3 validation set.
				AVG represents the average mAP at IoU thresholds 0.5:0.05:0.95.
				$*$ means fusing the detection results of the S-MIL model and our  P-MIL model.
				}
				\label{tab:activitynet1.3}
				\vspace{-0.5em}
				\begin{tabular}{c|cccc}
				\toprule
				\multirow{2.5}{*}{\bf Methods}                          & \multicolumn{4}{c}{\bf mAP@IoU (\%)}  \\
																	\cmidrule{2-5}
																	& 0.5           & 0.75          & 0.95         & AVG           \\
				\midrule
				STPN~\cite{cvpr2018stpn}, CVPR2018                  & 29.3          & 16.9          & 2.6          & 16.3          \\
				% STAR~\cite{aaai2019star}, AAAI2019                  & 31.1          & 18.8          & 4.7          & 18.2          \\
				CMCS~\cite{cvpr2019cmcs}, CVPR2019                  & 34.0          & 20.9          & 5.7          & 21.2          \\
				% ASSG~\cite{MM2019assg}, MM2019                      & 32.3          & 20.1          & 4.0          & 18.8          \\
				% TSM~\cite{iccv2019tsm}, ICCV2019                    & 30.3          & 19.0          & 4.5          & -             \\
				WSAL-BM~\cite{iccv2019backgroundmodel}, ICCV2019    & 36.4          & 19.2          & 2.9          & 19.5          \\
				% BaS-Net~\cite{aaai2020background}, AAAI2020         & 34.5          & 22.5          & 4.9          & 22.2          \\
				% ACM-BANet~\cite{mm2020acmnet}, MM2020               & 37.6          & 24.7          & \textbf{6.5} & 24.4          \\
				% A2CL-PT~\cite{eccv2020A2CL-PT}, ECCV2020            & 36.8          & 22.0          & 5.2          & 22.5          \\
				TSCN~\cite{eccv2020TSCN}, ECCV2020                  & 35.3          & 21.4          & 5.3          & 21.7          \\
				% WUM~\cite{aaai2021WSALUM}, AAAI2021                 & 37.0          & 23.9          & 5.7          & 23.7          \\
				TS-PCA~\cite{cvpr2021tspca}, CVPR2021               & 37.4          & 23.5          & 5.9          & 23.7          \\
				AUMN~\cite{cvpr2021aumn}, CVPR2021                  & 38.3          & 23.5          & 5.2          & 23.5          \\
				UGCT~\cite{cvpr2021ugct}, CVPR2021                  & 39.1          & 22.4          & 5.8          & 23.8          \\
				FAC-Net~\cite{iccv2021facnet}, ICCV2021             & 37.6          & 24.2          & 6.0          & 24.0          \\
				FTCL~\cite{cvpr2022ftcl}, CVPR2022                  & 40.0          & 24.3          & \textbf{6.4} & 24.8          \\
				RSKP~\cite{cvpr2022rskp}, CVPR2022                  & 40.6          & 24.6          & 5.9          & 25.0          \\
				ASM-Loc~\cite{cvpr2022asmloc}, CVPR2022             & 41.0          & 24.9          & 6.2          & 25.1          \\
				DCC~\cite{cvpr2022dcc}, CVPR2022                    & 38.8          & 24.2          & 5.7          & 24.3          \\
				\midrule
				% baseline                                            & 38.7          & 23.6          & 4.9          & 23.8          \\
				\textbf{ours} 	                                    & 39.5          & 23.6          & 4.9          & 23.9          \\
				\textbf{ours$^*$}  		                            & \textbf{41.8} & \textbf{25.4} & 5.2          & \textbf{25.5} \\
				\bottomrule
				\end{tabular}
				\vspace{-1em}
		\end{minipage}
	\end{minipage}
\end{figure*}

\subsection{Comparison with State-of-the-art Methods}
\label{sec:comparision with sota}
{\bf Results on THUMOS14}.
Table~\ref{tab:thumos14} shows the comparison of our method with weakly-supervised and several fully-supervised methods on the THUMOS14 dataset.
From the results we can see that our method outperforms the prior state-of-the-art weakly-supervised methods,
and by fusing the detection results of the S-MIL model and our P-MIL model,
we can even achieve better performance.
Specifically, our method surpasses the previous best performance by 1.5\% and 1.4\% in terms of the mAP@0.5 and the average mAP@0.1:0.7, respectively,
and further widens the gap to 1.7\% and 1.9\% after fusion.
Even when compared to certain fully-supervised methods (\emph{e.g.} BMN~\cite{iccv2019bmn} and GTAD~\cite{cvpr2020gtad}), our model can achieve comparable results at low IoU thresholds.

{\bf Results on ActivityNet}.
Table~\ref{tab:activitynet1.2} and Table~\ref{tab:activitynet1.3} show the performance comparison on the larger ActivityNet1.2 and ActivityNet1.3 datasets, respectively.
%
% Following the standard evaluation protocol~\cite{cvpr2015activitynet},
% we report the average mAP under IoU thresholds from 0.5 to 0.95 with a step size of 0.05.
%
The experimental results are consistent with those on the THUMOS14 dataset 
and we achieve state-of-the-art performance.
Specifically, after fusion, we achieve 26.5\% on the ActivityNet1.2 dataset and 25.5\% on the ActivityNet1.3 dataset in terms of the average mAP.

\begin{figure*}[t]
	% \vspace{-1em}
	\begin{minipage}[!t]{\linewidth}
	% \vspace{-0.8em}
	\centering
	\footnotesize
		\begin{minipage}[t]{0.48\linewidth}
			\makeatletter\def\@captype{table}
				\centering
				\caption{%
					Ablation studies about the candidate proposal generation methods.
					Action and background indicate that the candidate proposals are generated by $\theta_{act}$ and $\theta_{bkg}$, respectively.
				}
				\label{tab:proposal_generation}
				\vspace{-0.5em}
				\setlength{\tabcolsep}{7pt}{\begin{tabular}{p{10em}ccccc}
				\toprule
				\multicolumn{1}{c}{\multirow{2.5}{*}{\bf Proposal Generation}} & \multicolumn{5}{c}{\bf mAP@IoU (\%)}                                              \\
													 \cmidrule{2-6}
													 & 0.1           & 0.3           & 0.5           & 0.7           & AVG           \\
				\midrule
				\multicolumn{1}{c}{action}                               & 62.6          & 51.1          & 34.9          & 12.7          & 41.2          \\
				\multicolumn{1}{c}{action + background}                  & \textbf{70.9} & \textbf{57.8} & \textbf{39.8} & \textbf{14.4} & \textbf{46.5} \\
				\bottomrule
				\end{tabular}}

				\caption{%
					Ablation studies about different proposal scoring approaches.
					GT denotes using the IoU with the ground truth to score the candidate proposals.
					And FUSE means fusing the detection results of the S-MIL model and our P-MIL model.
				}
				\vspace{1em}
				\label{tab:proposal_scoring}
				\vspace{-0.7em}
				\setlength{\tabcolsep}{7pt}{\begin{tabular}{p{10em}ccccc}
				\toprule
				\multicolumn{1}{c}{\multirow{2.5}{*}{\bf Proposal Scoring}} & \multicolumn{5}{c}{\bf mAP@IoU (\%)}                                              \\
													\cmidrule{2-6}
													& 0.1           & 0.3           & 0.5           & 0.7           & AVG           \\
				\midrule
				\multicolumn{1}{c}{GT}                                & 83.3          & 76.3          & 64.7          & 41.7          & 67.4          \\
				\multicolumn{1}{c}{S-MIL}                             & 68.1          & 54.1          & 37.2          & 12.4          & 43.6          \\
				\multicolumn{1}{c}{P-MIL}                             & 70.9          & 57.8          & 39.8          & 14.4          & 46.5          \\
				\multicolumn{1}{c}{FUSE}                              & \textbf{71.8} & \textbf{58.9} & \textbf{40.0} & \textbf{15.1} & \textbf{47.0} \\
				\bottomrule
				\end{tabular}}
		\end{minipage}
		\hspace{3mm}
		\begin{minipage}[t]{0.48\linewidth}
			\makeatletter\def\@captype{table}
				\centering
				\caption{%
					Ablation studies about different variants of proposal feature extraction.
					The results demonstrate the effectiveness of the Surrounding Contrastive Feature Extraction module.
				}
				\label{tab:proposal_feature_extraction}
				\vspace{-0.5em}
				\setlength{\tabcolsep}{7pt}{\begin{tabular}{p{10em}ccccc}
				\toprule
				\multicolumn{1}{c}{\multirow{2.5}{*}{\bf \makecell[c]{Proposal Feature \\ Extraction}}} & \multicolumn{5}{c}{\bf mAP@IoU (\%)}                                              \\
															 \cmidrule{2-6}
															 & 0.1           & 0.3           & 0.5           & 0.7           & AVG           \\
				\midrule
				\multicolumn{1}{c}{w/o extending}                   & 64.5          & 52.0          & 34.3          & 11.1          & 41.0          \\
				\multicolumn{1}{c}{simply concatenate}                           & 65.8          & 53.1          & 35.1          & 11.2          & 41.9          \\
				\multicolumn{1}{c}{outer-inner contrast}                         & \textbf{70.9} & \textbf{57.8} & \textbf{39.8} & \textbf{14.4} & \textbf{46.5} \\
				\bottomrule
				\end{tabular}}

				\caption{%
					Ablation studies about the two designs of proposal refinement.
					PCE and IRC denote the Proposal Completeness Evaluation module and the Instance-level Rank Consistency loss, respectively.
				}
				\vspace{1em}
				\label{tab:proposal_refinement}
				\vspace{-0.5em}
				\setlength{\tabcolsep}{7pt}{\begin{tabular}{p{4em}p{4em}ccccc}
				\toprule
				\multicolumn{2}{c}{\bf Proposal Refinement} & \multicolumn{5}{c}{\bf mAP@IoU (\%)}                                              \\
				\midrule
				\multicolumn{1}{c}{PCE}                & \multicolumn{1}{c}{IRC}                & 0.1           & 0.3           & 0.5           & 0.7           & AVG           \\
				\midrule
													   &                    & 70.2          & 57.1          & 37.7          & 13.4          & 45.2          \\
				\multicolumn{1}{c}{\checkmark}         &                    & 70.4          & 57.6          & 38.7          & 14.5          & 45.9          \\
													   & \multicolumn{1}{c}{\checkmark}         & 70.6          & 58.0          & 39.0          & 13.8          & 46.0          \\
				\multicolumn{1}{c}{\checkmark}         & \multicolumn{1}{c}{\checkmark}         & \textbf{70.9} & \textbf{57.8} & \textbf{39.8} & \textbf{14.4} & \textbf{46.5} \\
				\bottomrule
				\end{tabular}}
		\end{minipage}
	\end{minipage}
	\vspace{-0.5em}
\end{figure*}

\begin{figure}[t!]
	\centering
	\includegraphics[width=0.95\linewidth]{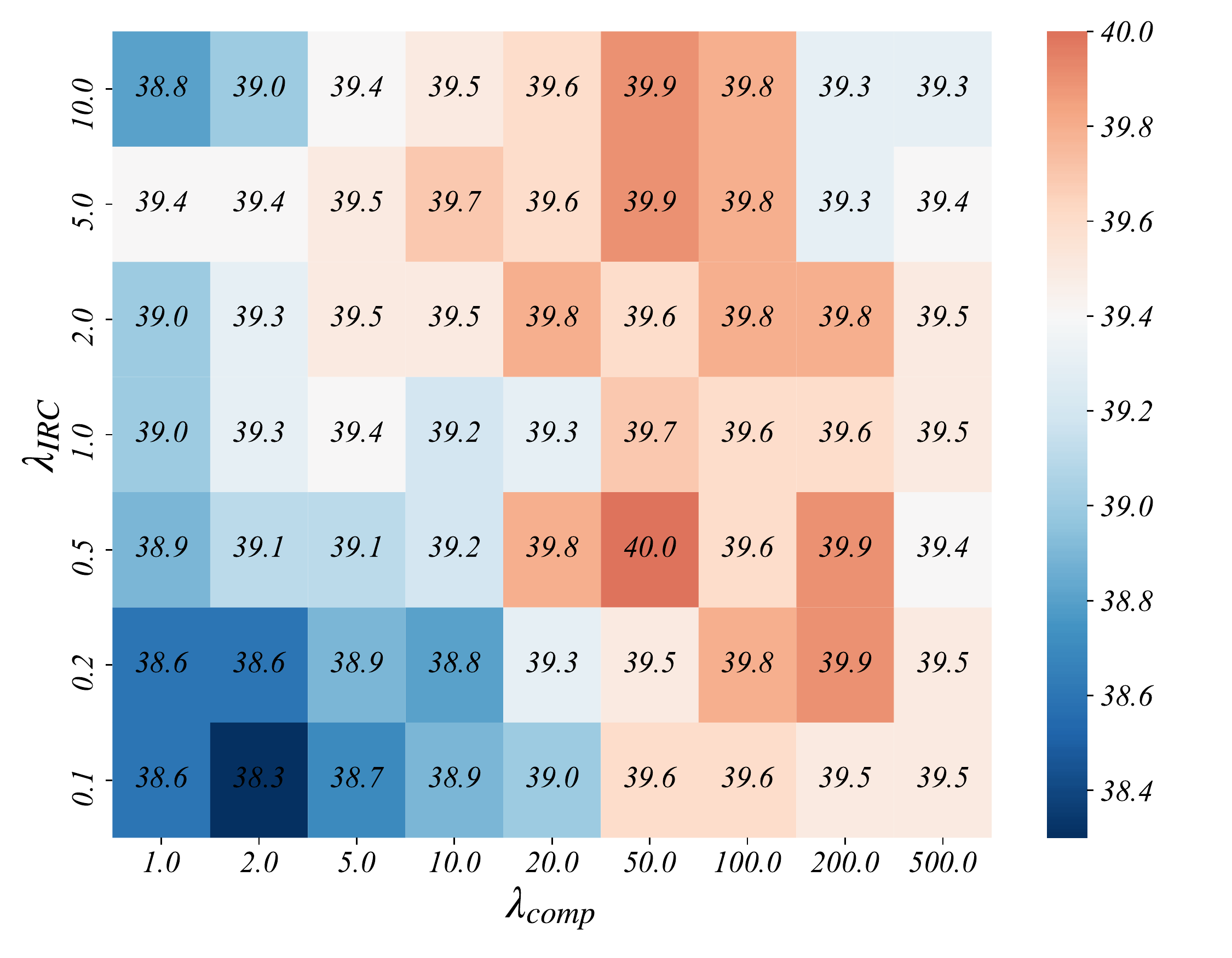}
	\vspace{-1em}
	\caption{
		The affection of the coefficients for the completeness loss and the Instance-level Rank Consistency (IRC) loss.
		We show the mAP at the IoU threshold 0.5.
	}
	\label{fig:ablation-weight}
	\vspace{-1em}
\end{figure}

\subsection{Ablation Studies}
\label{sec:ablation studies}
To analyze the impact of each design,
we conduct a series of ablation studies on the THUMOS14 dataset, as detailed below.

{\bf Proposal Generation}.
Table~\ref{tab:proposal_generation} shows the impact of different candidate proposal generation methods on the final detection performance.
To enable our P-MIL model to learn better foreground-background separation during training,
we generate additional background proposals to fill the candidate proposals in Equation~\eqref{eq:proposal}.
In order to validate the effectiveness of introducing the background proposals in the training stage,
we keep the candidate proposals used for testing to be consistent,
which consist of only the action proposals.
From Table~\ref{tab:proposal_generation}, we can see that when only the action proposals are used for training,
the average mAP is 41.2\%.
After incorporating the background proposals into the training stage,
the average mAP increases by 5.3\% to 46.5\%,
which significantly demonstrates the effectiveness of this design.

{\bf Proposal Scoring}.
Table~\ref{tab:proposal_scoring} shows the impact of different proposal scoring approaches on the detection performance.
To evaluate the upper bound of the detection performance, we use the IoU with the ground truth to score the candidate proposals.
% It can be observed that when the IoU with the ground truth is used to score the proposals, we can achieve a significantly high detection performance.
% reaches a considerable height.
The results indicate that the localization quality of the candidate proposals is already high enough
and the bottleneck of the detection performance lies in the scoring of the candidate proposals. 
To evaluate the effectiveness of our P-MIL method compared to the S-MIL method,
we apply different scoring approaches to the same set of candidate proposals.
It can be observed that when the S-MIL method is used to score the candidate proposals,
the average mAP is 43.6\%.
When we utilize our P-MIL method to score the candidate proposals,
the average mAP increases by 2.9\%.
The results show that the direct scoring of proposals by our P-MIL method is better than
the indirect scoring of proposals by the S-MIL method.
Note that after fusing the detection results of the S-MIL model and our P-MIL model,
the performance can be further improved to 47.0\% in terms of the average mAP,
indicating that the two methods can complement each other.

{\bf Proposal Feature Extraction}.
% {\bf Effectiveness of the Surrounding Contrastive Feature Extraction Module}.
%
Table~\ref{tab:proposal_feature_extraction} shows the impact of different variants of proposal feature extraction on the detection performance.
From the experimental results,
it can be seen that the average mAP is only 41.0\%
when the boundaries of the candidate proposals are not extended.
After extending the left and right boundaries of the candidate proposals,
we can obtain the feature vectors of the three regions.
However, simply concatenating these three feature vectors increases the average mAP by just 0.9\%.
When calculating the outer-inner contrastive features in Equation~\eqref{eq:contrast},
the performance is significantly improved by 5.5\% to 46.5\%.
These results validate the effectiveness of the Surrounding Contrastive Feature Extraction (SCFE) module.

{\bf Proposal Refinement}.
Table~\ref{tab:proposal_refinement} shows the impact of the two designs of proposal refinement on the detection performance,
including the Proposal Completeness Evaluation (PCE) module and the Instance-level Rank Consistency (IRC) loss.
It can be observed that both designs can bring performance gain.
Specifically, the PCE module and the IRC loss boost performance by 0.7\% and 0.8\% in terms of the average mAP, respectively,
and when used together, the performance increases by 1.3\%.
The experimental results demonstrate the effectiveness of both designs.

{\bf Hyper-parameters Sensitivity Analysis}.
There are two hyper-parameters in our P-MIL method,
including the coefficients $\lambda_{comp}$ and $\lambda_{IRC}$ of the loss function in Equation~\eqref{eq:prop_loss_total}.
To analyse the sensitivity of these hyper-parameters,
we evaluate the performance change in terms of the mAP@0.5 for different combinations of $\lambda_{comp}$ and $\lambda_{IRC}$.
As shown in Figure~\ref{fig:ablation-weight}, our model is not very sensitive to these two hyper-parameters, and the performance fluctuations in terms of mAP@0.5 are less than 2\%.
We set a moderate value for each of these two hyperparameters.
Specificly, with $\lambda_{comp}$ = 20 and $\lambda_{IRC}$ = 2, our method achieves 39.8\% in terms of the mAP@0.5.

% {\bf Error Analysis}.

\section{Conclusion}
\label{sec:conclusion}
In this paper, we propose a novel Proposal-based Multiple Instance Learning (P-MIL) framework for weakly-supervised temporal action localization, 
%
% {\color{red}
% Our core insight is that the bottleneck of the WTAL methods lies in the scoring of the proposals
% and directly scoring the proposals is more effective than indirectly aggregating the segment-level predictions.}
which can achieve the unified objectives of the training and testing stages by directly classifying the candidate proposals. 
We introduce three key designs to deal with the challenges in different stages of the P-MIL framework,
including the surrounding contrastive feature extraction module, the proposal completeness evaluation module and the instance-level rank consistency loss.
Extensive experimental results on two challenging benchmarks demonstrate the effectiveness of our method.

\section{Acknowledgement}
\label{sec:acknowledgement}
This work was partially supported by the National Nature Science Foundation of China (Grant 62022078, 62121002), and National Defense Basic Scientific Research Program of China (Grant JCKY2021130B016).

% \appendix
% \input{supplementary}

%%%%%%%%% REFERENCES
{\small
\bibliographystyle{ieee_fullname}
\bibliography{wsal}
}

\end{document}